\documentclass[letterpaper, 10 pt, conference]{ieeeconf}  
\IEEEoverridecommandlockouts                              
\let\labelindent\relax
\usepackage{enumitem}
\usepackage{xspace}
\usepackage{xcolor,soul}
\usepackage{graphicx}
\usepackage{subcaption}
\usepackage{graphicx}
\usepackage{tabularx}
\usepackage{booktabs}
\usepackage{multirow}
\usepackage[accsupp]{axessibility}  
\usepackage[pagebackref,breaklinks,colorlinks,citecolor=eccvblue]{hyperref}
\usepackage{orcidlink}
\usepackage{siunitx}
\usepackage{breqn}

\usepackage{tikz}
\usetikzlibrary{arrows.meta, positioning, fit}

\newcommand{\names}{GVT3D\xspace}

\newcommand{\name}{GVT3D}
\newcommand{\namef}{GVT3D's\xspace}
\newcommand{\dataset}{ImputeRefer\xspace}

\usepackage{graphicx} 

\title{\name: Gesture-Vision-Text Token Enrichment for 3D Embodied Reference Understanding} 

\author{%
Vimarsh Shah$^{1}$, Dulanga Weerakoon$^{2}$, Ajinkya Deshpande$^{1}$, Saksham Agarwal$^{1}$,\\
Nilesh Bhatia$^{1}$, Neel Kiran Naik$^{1}$, Parth Shah$^{1}$, Sravan Danda$^{1}$, Sanjay E. Sarma$^{3}$,\\
Archan Misra$^{4}$, Vigneshwaran Subbaraju$^{5,6}$, Sougata Sen$^{1}$%
\thanks{$^{1}$BITS Pilani Goa campus, India. Emails: Vimarsh Shah: \texttt{f20221060@goa.bits-pilani.ac.in}; Ajinkya Deshpande: \texttt{f20220746@goa.bits-pilani.ac.in}; Saksham Agarwal: \texttt{f20230521@goa.bits-pilani.ac.in}; Nilesh Bhatia: \texttt{f20230418@goa.bits-pilani.ac.in}; Neel Kiran Naik: \texttt{f20230429@goa.bits-pilani.ac.in}; Parth Shah: \texttt{f20221168@goa.bits-pilani.ac.in}; Sravan Danda: \texttt{dandas@goa.bits-pilani.ac.in}; Sougata Sen: \texttt{sougatas@goa.bits-pilani.ac.in}.}%
\thanks{$^{2}$Singapore-MIT Alliance for Research and Technology (SMART) Centre, Singapore. Email: Dulanga Weerakoon: \texttt{dulanga.weerakoon@smart.mit.edu}.}%
\thanks{$^{3}$Massachusetts Institute of Technology, USA. Email: Sanjay E. Sarma: \texttt{sesarma@mit.edu}.}%
\thanks{$^{4}$Singapore Management University, Singapore. Email: Archan Misra: \texttt{archanm@smu.edu.sg}.}%
\thanks{$^{5}$Agency for Science, Technology and Research (A*STAR), Singapore. Email: Vigneshwaran Subbaraju: \texttt{Vigneshwaran\_Subbaraju@a-star.edu.sg}.}%
\thanks{$^{6}$CNRS@CREATE, Singapore.}%
}

\begin{document}

\maketitle

\begin{abstract}
Natural human–robot interaction requires robots to interpret language and gestures within a 3D environment to support tasks such as pick-and-place and object retrieval. 3D Embodied Reference Understanding (3D-ERU) supports this by identifying the 3D location of a target object from a scene (point-cloud) containing a human pointing gesture and an accompanying language description. However, current 3D-ERU models lack alignment of cross-modal representations and incur high processing costs that increase with the point-cloud's size. To address this, we propose \emph{\names}, which employs an Embodied Fusion module that jointly models language, gesture, and vision inputs within a shared embedding space, promoting cross-modal alignment. We additionally introduce a foveation approach that selectively processes the point cloud within a coarse region around the estimated pointing direction to reduce computations. \names achieves $\sim$8\% higher accuracy and foveation enables 2.95$\times$ faster execution while maintaining comparable accuracy, relative to prior 3D-ERU models.
\end{abstract}



\section{Introduction}\label{sec:intro}
Fluent, natural interaction between humans and robots requires a robot to accurately interpret the natural linguistic and gestural expressions generated concurrently in natural human communication~\cite{kendon2004gesture}. In this context, an \emph{embodied reference} denotes a human instruction comprising a language expression accompanied by an optional pointing gesture that refers to a single target object within a scene~\cite{chen2021yourefit}; such multimodal instructions are common for \emph{pick-and-place} or \emph{object retrieval} robotic tasks associated with home and warehouse logistics~(e.g.,~\cite{ji2024foundation}), as presented in Figure~\ref{fig:introFig}. The capability to accurately resolve such commands, through joint reasoning over language and gesture signals, is referred to as `embodied reference understanding'.


Embodied reference understanding has been explored using deep neural network-based approaches that operate on 2D visual inputs captured from ego-centric and exo-centric views \cite{chen2021yourefit,COSM2IC}. 3D Embodied Reference Understanding (3D-ERU) extends this problem into identifying a target object within a \textit{3D scene}, typically represented as a point cloud captured from a global, exo-centric view via LiDARs installed in the infrastructure \cite{Lu2024ScanERU, Mane2025Ges3Vig}. 
Compared to 2D settings, this introduces additional challenges due to spatial depth reasoning, geometric ambiguity, and articulated human pose variability, but has remained relatively underexplored; Ges3ViG~\cite{Mane2025Ges3Vig} and ScanERU~\cite{Lu2024ScanERU} being among the few approaches that address this problem in 3D settings. 
The limited literature on 3D-ERU leaves the following gaps unexplored:

\begin{figure}
    \centering
    \begin{subfigure}[b]{0.4\linewidth}
        \centering
        \includegraphics[width=\linewidth]{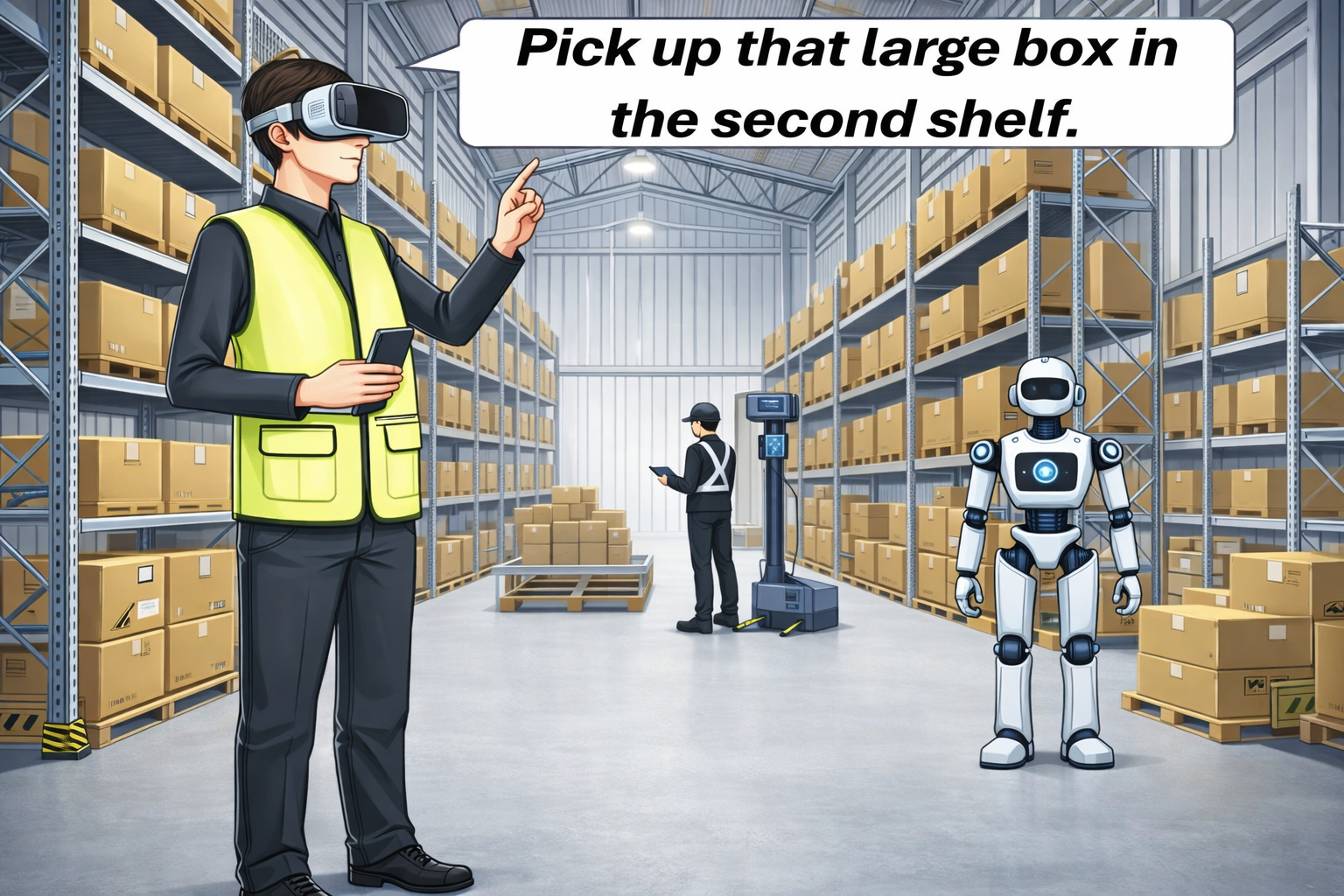}
        \caption{}
    \end{subfigure}
    \hfill
    \begin{subfigure}[b]{0.2\linewidth}
        \centering
        \includegraphics[width=0.42\linewidth]{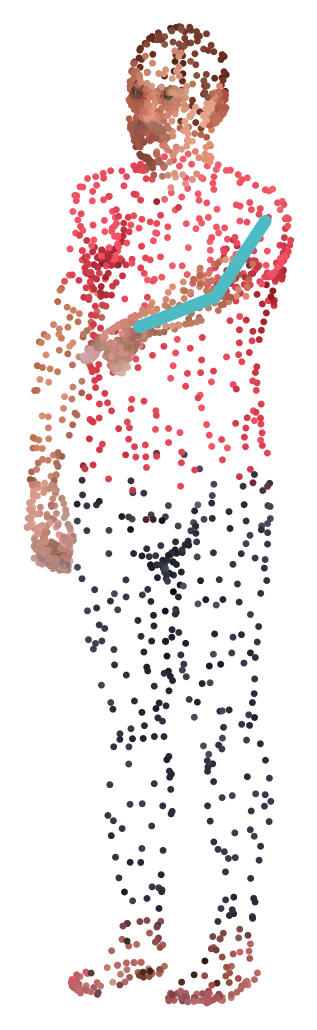}
        \caption{}
    \end{subfigure}
        \hfill
    \begin{subfigure}[b]{0.36\linewidth}
        \centering
        \includegraphics[width=\linewidth]{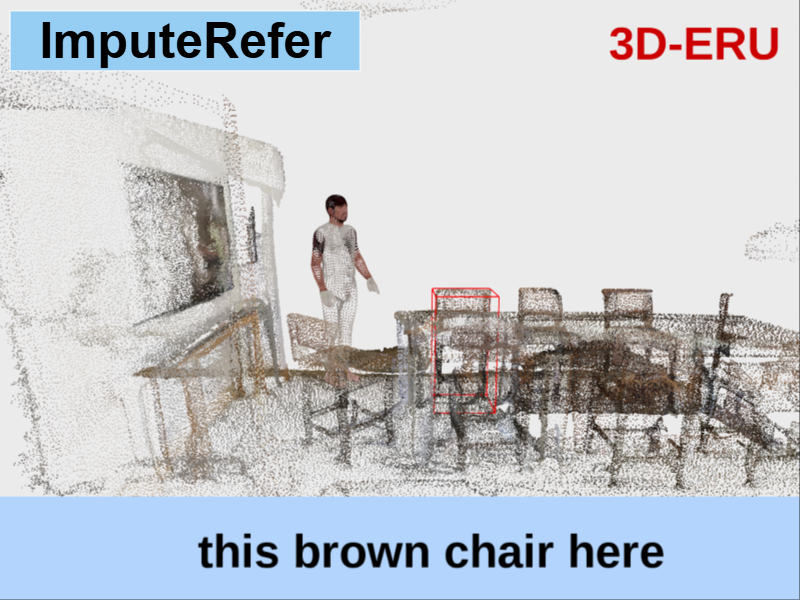}
        \caption{}
    \end{subfigure}
    \caption{(a) A practical task requiring 3D-ERU through the fusion of gesture, vision and text. (b)~Point-cloud of a human performing gesture. (c)~A sample of 3D-ERU from Imputerefer dataset. 
    }
    \label{fig:introFig}
    \vspace{-0.25in}
\end{figure}

 
 \begin{itemize}[leftmargin=*]
    \item \emph{Lack of Unified Cross-Modal Semantics:} Despite substantial progress in grounding linguistic referring expressions, existing methods for 3D-ERU do not explicitly model the semantics of interaction between language and gesture within a common representation space. Prior work on 3D-ERU incorporates gesture only through late fusion, where the estimated pointing direction is used as the final arbiter to re-weight or select candidate regions along the pointing ray, e.g., Ges3ViG~\cite{Mane2025Ges3Vig}. Such formulations treat gesture as an auxiliary signal operating independently of language, rather than truly exploiting the interactions between language and gesture that  jointly constrain the identification and localization of the target. Previous studies have also shown that relying solely on pointing can be inaccurate due to systematic human pointing biases and variability in articulation~\cite{li2023understanding}, and that the relative importance of language and pointing may depend on varying environmental context~\cite{weerakoon2020gesture}. 
    Thus, gestures must be interpreted jointly with language throughout the reasoning process, with the relative importance adjusted based on factors like pointing fidelity, or presence of confounding distractor objects.
    
    \item \emph{Lack of Computational Efficiency:} Existing models for 3D-ERU \cite{Lu2024ScanERU,Mane2025Ges3Vig} remain untested in practical real-world scenarios involving downstream tasks/reasoning on top of 3D-ERU being executed \emph{on-device by a robot}: 
    for example, enabling a robot to rapidly and accurately navigate to the object specified by an embodied reference. Extant 3D-ERU models incur substantial computational overhead that scales with the number of points; we show in Section~\ref{subsec:accLatEne} that this prohibitive overhead can meaningfully affect robot responsiveness on embedded edge computing platforms.
    Consequently, 2D ERU models~\cite{COSM2IC,weerakoon2020gesture} remain widely preferred and adopted, even though 2D reasoning limits the ability to identify the 3D coordinates of the target object in the real world. Computationally efficient, \emph{edge-compatible} 3D-ERU models are thus needed to increase adoption in practical deployments.
\end{itemize} 

In this paper, we introduce \emph{\textbf{G}}esture-\textbf{\emph{V}}ision-\textbf{\emph{T}}ext Token Enrichment for \emph{\textbf{3D}} Embodied Reference Understanding (\emph{\textbf{\names}}), a new model that explicitly aligns gesture, vision, and language features within a shared embedding space through a novel embodied fusion mechanism, enabling semantically-rich structured cross-modal interaction 
before any modality-specific score is computed. 
Rather than relying solely on late fusion or simple feature concatenation, \names combines this joint representation learning with a final learned re-weighting stage, allowing the model to adaptively trust language, gesture, or vision depending on their relative reliability for a given instance, leading to improved accuracy compared to baseline 3D-ERU models. 
To further address the computational cost of processing large 3D scenes, we propose point cloud \emph{{foveation}}, 
whereby candidate objects are first ranked by how well they align with the pointing direction determined via a lightweight gesture predictor, and the expensive 2D visual encoding step is executed only on the most gesture-consistent proposals. 
In summary, we make the following \textbf{key contributions:}

\begin{enumerate}[leftmargin=*]
\item \textbf{GVT3D model for 3D-ERU:} 
On the ImputeRefer~\cite{Mane2025Ges3Vig} benchmark, our \names outperforms Ges3ViG (prior SOTA) by achieving \textbf{8}\% higher accuracy, alongside a 1.75$\times$ speedup. Our analysis further shows that \names meaningfully aligns the vision, language and gesture embeddings, enabling it to adjust the importance of individual modality for a given input context.


\item \textbf{Foveation for Efficient 3D-ERU with \name:} To address the computational overhead of processing large point clouds found in 3D-ERU data, we propose point cloud foveation-based computation. 
Compared to Ges3ViG, \names with foveation runs \textbf{2.95$\times$} faster and reduces energy consumption by 70.76\% while improving accuracy by 2.75\%.

\item \textbf{Real-World Prototype of \names with Foveation:} We implement a robotic system that integrates \names as its reasoning module and Nav2~\cite{macenski2023survey} for navigation planning to support a representative \emph{object retrieval} task. 
In real-world evaluations, foveation-enhanced \names achieves a success rate of {70\%} across 30 scenes which the model hasn't seen before, in comparison to 56\% achieved by Ges3ViG on the same task.
\end{enumerate}

\section{Background and Related Work}\label{sec:related}

\subsection{2D Visual Grounding}

Localising image regions from natural language descriptions, commonly known as Referring Expression Comprehension, has been a long-standing challenge in vision-language understanding. Early approaches followed a proposal-based paradigm where candidate regions were generated and matched with language features using multimodal embeddings~\cite{hu2016natural}. More recent work has explored transformer-based architectures that align textual tokens with spatial regions through attention mechanisms~\cite{Deng_2021_ICCV, kamath2021mdetr}. In parallel, large-scale vision-language pretraining has enabled strong semantic alignment between images and text~\cite{radford2021learning}. Despite these advances, most 2D grounding methods assume that language uniquely specifies the referent, ignoring the use of additional cues, such as gestures and gaze, in natural human-robot communication~\cite{jia2026chi}.

\subsection{2D-Embodied Reference Understanding}

Recent work extends referring-expression comprehension by incorporating embodied cues such as pointing gestures alongside language. 
In the 2D setting, this integration has largely focused on system-level efficiency rather than representation-level fusion. 
For instance,
COSM2IC enhances real-time multimodal instruction understanding~\cite{COSM2IC}, involving 2D vision, language and pointing gestures, by introducing adaptive model selection to balance accuracy and computational efficiency, rather than jointly modeling how gesture and language features interact. 

\subsection{3D Visual Grounding}

3D visual grounding extends referring-expression comprehension to 3D scene representations, typically point clouds, requiring reasoning over spatial geometry and object layouts. Early work, such as ScanRefer, introduced large-scale datasets and proposal-based grounding pipelines built upon ScanNet scenes~\cite{dai2017scannet, chen2020scanrefer}. Subsequent work, including ReferIt3D, highlighted the challenges of spatial reasoning in cluttered environments~\cite{achlioptas2020referit3d}. More recent methods incorporate attention mechanisms to associate language tokens with object-level geometric features~\cite{yuan2021instancerefer}, or adopt transformer-based architectures for end-to-end grounding~\cite{he2021transrefer3d}. Leveraging vision-language pretraining has further improved performance by aligning rendered multi-view images with textual embeddings~\cite{m3drefclip}, while approaches such as PointCLIP adapt CLIP representations for point cloud understanding through multi-view projections~\cite{zhang2021pointclip}.

\subsection{3D-Embodied Reference Understanding}

3D embodied reference understanding studies how agents interpret multimodal instructions that combine language with physical cues like pointing gestures in 3D environments. 
Ges3ViG originally introduced approaches to incorporate pointing gestures into 3D visual grounding by modeling gesture as a geometric ray extending from the shoulder to fingertip~\cite{Mane2025Ges3Vig}. 
ScanERU further investigates interactive grounding using human pose signals to derive spatial constraints~\cite{Lu2024ScanERU}. 
However, these methods typically represent gestures using deterministic geometric formulations and integrate gesture information through late fusion strategies.   In contrast, our work treats gesture, vision, and language as tightly coupled modalities and jointly models the semantics of their interactions through a unified token-level fusion framework.
\begin{figure*}[t]
    \centering
    \includegraphics[width=0.95\linewidth]{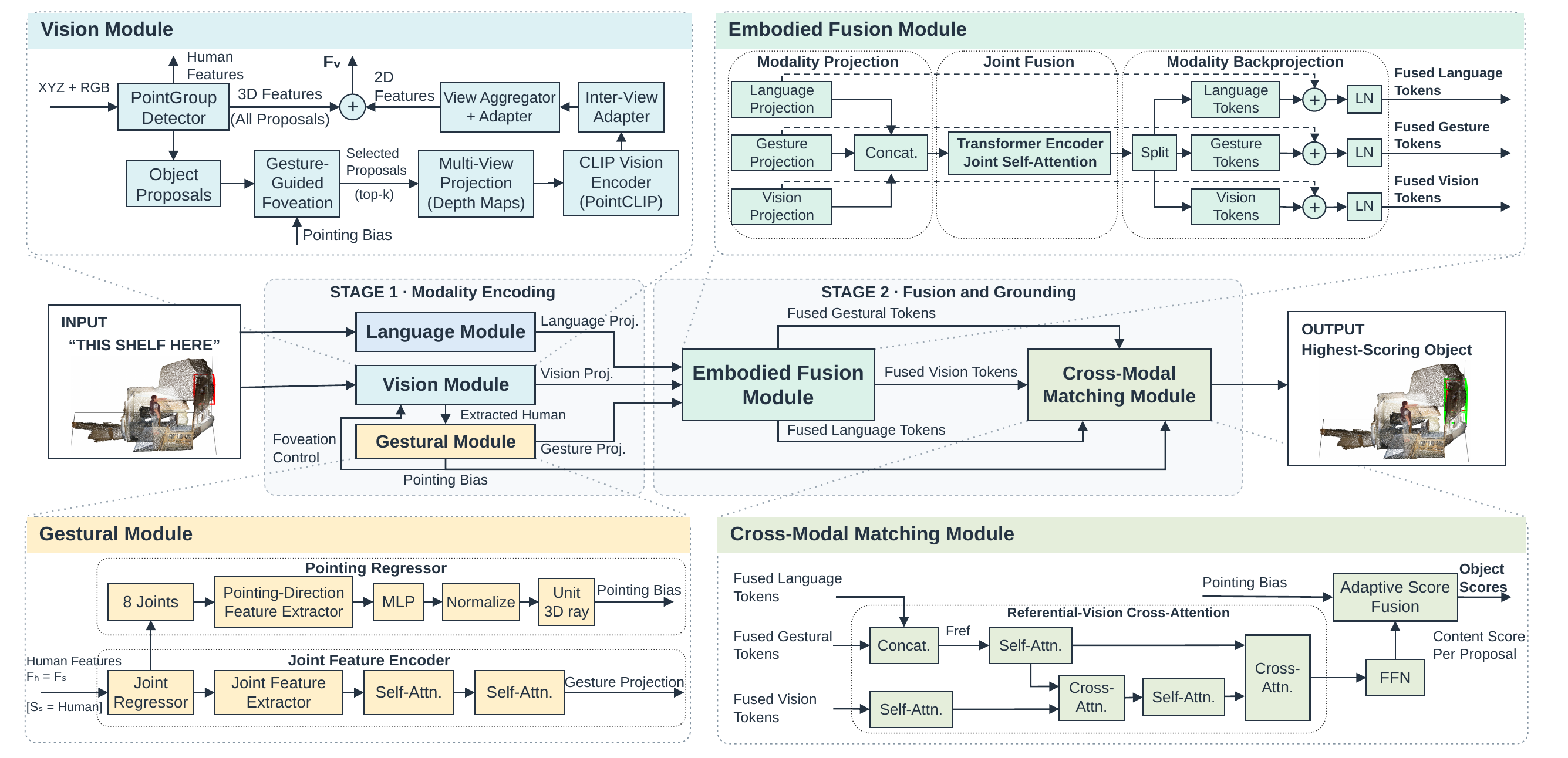}
    \caption{Proposed \names Approach }
    \label{fig:gvt3d_model_arch_diag_with_stages}\vspace{-0.25in}
\end{figure*}

\section{\names: Gesture-Vision-Text for 3D-ERU}
\label{sec:gvt3d}









\label{sec:method}


We now present the proposed \names approach, illustrated in Figure~\ref{fig:gvt3d_model_arch_diag_with_stages}. 
\names differs from prior 3D-ERU methods in two ways. 
(i)~we estimate the pointing direction using a learned model instead tracing a shoulder-to-fingertip line, to capture natural arm gesture. 
(ii)~instead of using gesture as a separate signal for refining object selection, \namef  Embodied Fusion Module performs joint reasoning over gesture, language, and vision tokens, allowing each modality to inform others before the target object is chosen.


\subsection{Stage 1: Per Modality Encoding}
The first stage in the \names architecture encodes the three signal streams: vision, language, and gesture. 

\subsubsection{Vision Module}
The vision module encodes the scene point cloud via a PointGroup detector~\cite{pointgroup}, which takes a 3D point cloud $\mathbf{P} \in \mathbb{R}^{N \times C}$ as input, where $N$ is the number of points and $C = 6$ (three coordinates and three RGB values). The detector outputs: (i) per-point semantic features $\mathbf{F}_S \in \mathbb{R}^{N \times d_{\text{det}}}$ and scores $\mathbf{S}_S \in \mathbb{R}^{N \times N_c}$, used for gesture inference; and (ii) proposal features $\mathbf{F}_p^{\text{3D}} \in \mathbb{R}^{M \times d_{\text{det}}}$ and objectness scores $\mathbf{S}_p \in \mathbb{R}^{M}$, used for proposal ranking, where $d_{\text{det}} = 32$, $M$ is the number of proposals, and $N_c$ is the number of semantic classes. To selectively allocate computation in the complementary 2D branch, we introduce Gesture-Guided Foveation before multi-view depth projection. This selects a subset of proposals for depth projection and CLIP encoding, while preserving $\mathbf{F}_p^{\text{3D}}$ for all $M$ proposals.



\paragraph{Gesture-Guided Foveation}
We use the unit pointing direction
$\hat{\mathbf{d}}_{\text{pred}}$ estimated by the Pointing
Regressor in the Gestural Module, described in
Section~\ref{sec:gesture_module_method}, to select proposals for
2D visual encoding. 
For proposal $i$, let $\mathbf{c}_i$ denote its bounding-box
centre and $\mathbf{a}$ denote the shoulder joint of the pointing arm. 
Its pointing bias is
\begin{equation}
    b_i =
    \frac{
        (\mathbf{c}_i-\mathbf{a})^\top \hat{\mathbf{d}}_{\text{pred}}
    }{
        \max\!\left(
            \|\mathbf{c}_i-\mathbf{a}\|_2,\epsilon
        \right)
    }
    \label{eqn:Pointing_Bias}
\end{equation}
where $\epsilon$ ensures numerical stability. This is the same
pointing bias used subsequently in Adaptive Score Fusion.

When foveation is enabled and a human is detected, proposals are
ranked by $b_i$, and the highest-scoring $K_f$ number of proposals are selected where $K_f=\min\!\left(M,\max(k,k_{\min})\right)$. Here, $k$ is the configured proposal budget and $k_{\min}$ is the
minimum number of proposals to retain for 2D encoding. Let $\mathcal{I}_f$ contains the indices of the selected proposals.
We construct the
binary selection mask $m_i=1[i\in\mathcal{I}_f]$ and perform
depth projection and CLIP encoding only for proposals
for which $m_i=1$. Only selected proposals undergo depth projection (described later) and CLIP encoding. This proposal-level foveation preserves the spatial
resolution of each generated depth map, while reducing
number of proposals processed by the 2D branch. 

\paragraph{Multi-view depth projection}
For each of the selected proposals, the scene point cloud is cropped using the predicted axis-aligned bounding box, centred at the origin, and normalised to a unit sphere. 
The normalised cloud is then projected via orthographic projection onto $V = 3$ single-channel depth maps of size $224 \times 224$, with viewpoints uniformly distributed over the sphere. 
Each pixel stores the normalised depth of the nearest projected point; RGB information is not used.

\paragraph{CLIP visual encoding}
Each depth map is replicated across three channels and normalised using the standard CLIP image statistics. Depth images from selected, nonempty proposals are encoded in a single forward pass through the frozen CLIP ViT-B/32 visual backbone, yielding per-view features $\mathbf{F}_{\text{view}} \in \mathbb{R}^{M \times V \times d_{\text{clip}}}$, where $d_{\text{clip}} = 512$. This approach, inspired by PointCLIP~\cite{zhang2021pointclip}, operates directly on raw 3D geometry rather than rendered RGB images.

\paragraph{Inter-view adapter}
To fuse information across the $V$ viewpoints of proposal $i$, we introduce a bottleneck MLP with a residual connection. For each proposal, the $V$ view vectors are concatenated along the feature dimension and passed through the MLP:
\begin{equation}
    \mathbf{F}_{\text{adapt,$i$}}
    = \mathbf{F}_{\text{view,$i$}}
    + \operatorname{reshape}\!\Bigl(
        \operatorname{MLP}_{\text{bn}}\!\bigl(
            \operatorname{flatten}(\mathbf{F}_{\text{view,$i$}})
        \bigr),\;V,\; d_{\text{clip}}
    \Bigr)
\end{equation}
where $\operatorname{flatten}(\mathbf{F}_{\text{view,$i$}}) \in \mathbb{R}^{V \cdot d_{\text{clip}}}$ and $\operatorname{MLP}_{\text{bn}}$ is a two-layer MLP that reduces dimensionality by a factor of 4 before projecting back to $V \cdot d_{\text{clip}}$.

\paragraph{View aggregator and projector}
The adapted view features are average-pooled across the view dimension and passed through a two-layer projection MLP to produce proposal-level 2D features:
\[
\mathbf{F}^{2D}_{p,i}
=
\operatorname{MLP}\!\left(\frac{1}{V}\sum_{v=1}^{V}\mathbf{F}^{(v)}_{\text{adapt,$i$}}\right)
\in \mathbb{R}^{M \times d_{\text{vis}}},
\qquad d_{\text{vis}}=128.
\]

\paragraph{Combining visual features}
For proposals not selected by foveation ($m_i = 0$), $F_p^{2D}$ is set to a zero vector. 
The 3D proposal features from PointGroup and the 2D features from the PointCLIP pipeline are concatenated and L2-normalised to form the joint visual representation:
\begin{equation}
    \mathbf{F}_v
    = \ell_2\!\bigl([\,\mathbf{F}_p^{\text{3D}} \;\|\; \mathbf{F}_p^{\text{2D}}\,]\bigr)
    \in \mathbb{R}^{M \times (d_{\text{det}} + d_{\text{vis}})}
    = \mathbb{R}^{M \times 160},
\end{equation}
where $[\cdot \| \cdot]$ denotes feature-dimension concatenation and $\ell_2(\cdot)$ denotes L2-normalisation. The resulting representation retains all $M$ proposals for further processing.

\subsubsection{Language Module}

We adopt the CLIP based language encoder, replicating the approach adopted in Ges3ViG~\cite{Mane2025Ges3Vig}. 
The referring expression is tokenized using the CLIP tokenizer, and both token-level and sentence-level representations are extracted from the frozen CLIP text transformer. The token level features are projected through a two-layer MLP to a shared embedding dimension $d_t = 128$, yielding word-level features $\mathbf{F}_{l,t} \in \mathbb{R}^{N{\text{tok}} \times 128}$. The sentence level representation is obtained via a learned projection head applied to the global CLIP text embedding, resulting in $\mathbf{F}_{l,s} \in \mathbb{R}^{128}$.



\subsubsection{Gestural Module} \label{sec:gesture_module_method}
In contrast to work like Ges3ViG~\cite{Mane2025Ges3Vig}, which models pointing gestures using a heuristic shoulder-to-fingertip ray, we introduce a learned pointing direction estimator. 
In natural human interactions, pointing gestures are rarely perfectly straight arms~\cite{li2023understanding}, so a fixed geometric ray does not always reflect the intended target direction. 
To address this, we propose a gestural module that learns the pointing direction directly from joint positions, rather than assuming a fixed arm geometry.

\paragraph{Joint Feature Encoder}
This module extracts the human point cloud segment identified by the PointGroup detector predictions. 
Using per-point semantic features $\mathbf{F}_s \in \mathbb{R}^{N \times 32}$ and semantic scores, points classified as \texttt{human} are selected and fed through a PointNet-style joint regressor that predicts soft assignment weights, $\mathbf{W}_{\text{hum}} \in \mathbb{R}^{N_{\text{hum}} \times N_j}$ for $N_j\!=\!55$ SMPL-X joints~\cite{pavlakos2019expressive}. 
Joint 3D coordinates are computed as $\mathbf{J} = \text{Softmax}(\mathbf{W}_{\text{hum}}^T) \times \mathbf{P}_{\text{hum}}$ and initial joint features as $\mathbf{F}_{j,\text{init}} = \text{Softmax}(\mathbf{W}_{\text{hum}}^T) \times \mathbf{F}_{s,\text{hum}}$, similar to computation in Point2Skeleton~\cite{Lin_2021_CVPR_point2skeleton}. 
These initial features are processed through a \emph{Joint Feature Extraction} module consisting of 1D convolution layers ($32 \!\to\! 64 \!\to\! 128$) with LeakyReLU activations, producing joint-level features $\mathbf{F}_{j} \in \mathbb{R}^{N_j \times 128}$. 
The module also computes a global aggregate feature $F_{agg}$ by average pooling intermediate representations at each layer and concatenating them. 
This aggregate feature captures the global pose information. 

The Joint Feature Extractor's output is sequentially passed through two self-attention layers, each comprising a multi-head attention mechanism ($h\!=\!4$, $d_k\!=\!d_v\!=\!32$). This outputs a set of gesture tokens $\mathbf{G} \in \mathbb{R}^{B \times N_j \times 128}$ that encodes pose information for the downstream Joint Fusion task. 
\paragraph{Pointing Regressor}

With the computed joint features, we employ a lightweight MLP termed the Pointing Regressor, which takes as input the 3D positions of eight key joints from a set of 55 total human joints, namely the shoulder, elbow, wrist, and index fingertip for both left and right arms. For each arm, we compute shoulder-centred relative offsets:\vspace{-0.15in}
\begin{dmath}
    \boldsymbol{\delta}_{\text{arm}} = \bigl[
        \mathbf{p}_{\text{elbow}} - \mathbf{p}_{\text{shoulder}},\;
        \mathbf{p}_{\text{wrist}} - \mathbf{p}_{\text{shoulder}},\;
        \mathbf{p}_{\text{index}} - \mathbf{p}_{\text{shoulder}},\;
        {\mathbf{p}_{\text{wrist}} - \mathbf{p}_{\text{elbow}}\bigr]~\in~\mathbb{R}^{12}}
\end{dmath}
The offset vectors from both arms are concatenated to form a 24-dimensional input, which is then passed through a three-layer MLP ($24 \!\to\! 64 \!\to\! 64 \!\to\! 3$) with ReLU activations. 
The output is L2-normalised to produce the predicted unit pointing direction $\hat{\mathbf{d}}_{\text{pred}}$. 
This $\hat{\mathbf{d}}_{\text{pred}}$ is used to compute proposal-wise
pointing biases, which guide top-$k$ proposal selection
for foveated visual encoding and contribute to adaptive
score fusion in the Cross-Modal Matching Module.

The inclusion of $\mathbf{p}_{\text{wrist}} - \mathbf{p}_{\text{elbow}}$ explicitly captures the forearm direction independently of the upper arm. 
This allows the regressor to learn the relationship between joint geometry and pointing direction directly from the data, rather than assuming a fixed shoulder-to-fingertip line. 




\begin{figure*}[t]
    \centering
    \includegraphics[width=1.9\columnwidth]{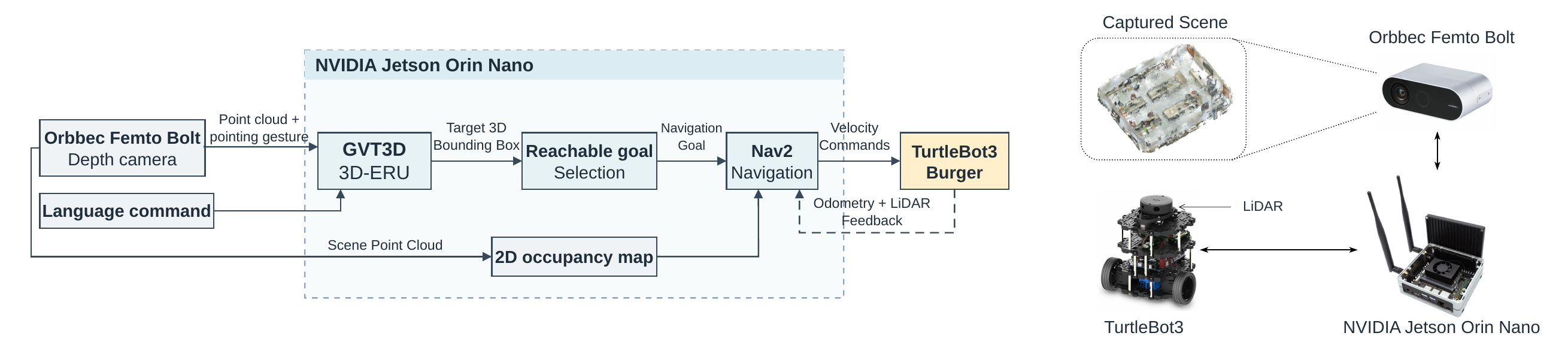}
    \caption{Robotic Navigation Implementation}
    \label{fig:nav_implementation_architecture}
    \vspace{-0.2in}
\end{figure*}

\subsection{Stage 2: Fusion}
Next, the projections from the three primary modules in Stage 1 are fused in Stage 2, which we describe next. 

\subsubsection{Embodied Fusion Module} \label{3wayblock}

Prior work on 3D-ERU incorporates gestural features using late fusion strategies, where each modality is processed largely independently and combined only at the final stages~\cite{Mane2025Ges3Vig}. 
Instead, we formalise gesture, vision, and text as three tightly coupled modalities that should be projected into a shared embedding space and conditioned on one another throughout representation learning. 

We propose a novel feature fusion mechanism termed \emph{{Embodied Fusion Module}}. 
Concretely, we design a \textit{3-way joint self attention} computation that enables mutual conditioning across gesture, vision, and text tokens, resulting in enriched modality representations that are better aligned for downstream grounding.
We now detail the components of our embodied fusion module: 

\paragraph{Modality Projection Layer} 
Each modality has a different native dimensionality: text tokens lie in $\mathbb{R}^{d_t}$ ($d_t\!=\!128$), gesture tokens in $\mathbb{R}^{d_g}$ ($d_g\!=\!128$), and vision tokens in $\mathbb{R}^{d_v}$ ($d_v\!=\!160$). We project each into a common fusion space of dimension $d_f\!=\!256$ via learned bias-free linear layers as shown below:
\begin{align}
    \mathbf{T}_f &= \mathbf{F}_{l,t}\, \mathbf{W}_t, \quad \mathbf{W}_t \in \mathbb{R}^{d_t \times d_f}, \\
    \mathbf{G}_f &= \mathbf{G}\, \mathbf{W}_g, \quad\;\; \mathbf{W}_g \in \mathbb{R}^{d_g \times d_f}, \\
    \mathbf{V}_f &= \mathbf{F}_v\, \mathbf{W}_v, \quad\;\; \mathbf{W}_v \in \mathbb{R}^{d_v \times d_f}.
\end{align}

\paragraph{Joint Fusion Layer}
The projected tokens are concatenated along the sequence dimension into a single sequence of length $L_t + N_j + K$:
\begin{equation}
    \mathbf{S} = [\mathbf{T}_f;\; \mathbf{G}_f;\; \mathbf{V}_f] \in \mathbb{R}^{B \times (L_t + N_j + K) \times d_f}.
\end{equation}
The concatenated sequence is processed by a Transformer encoder layer~\cite{vaswani2023attentionneed} ($h=8$ attention heads, FFN dimension $4d_f$). 
Self-attention over the full sequence enables cross-modal interaction among all tokens.

\paragraph{Modality Backprojection Layer} 
Next, the output sequence is partitioned into the original text, gesture, and vision segments. 
Each segment is projected back to its native feature dimension using a modality-specific linear layer, followed by a residual connection and layer normalisation: \vspace{-0.1in}
\begin{align}
    \mathbf{F}'_{l,t} &= \text{LN}_{t}\!\bigl(\mathbf{F}_{l,t} + \mathbf{W}'_t\, \mathbf{T}'_f\bigr), \\
    \mathbf{G}' &= \text{LN}_{g}\!\bigl(\mathbf{G} + \mathbf{W}'_g\, \mathbf{G}'_f\bigr), \\
    \mathbf{F}'_v &= \text{LN}_{v}\!\bigl(\mathbf{F}_v + \mathbf{W}'_v\, \mathbf{V}'_f\bigr).
\end{align}


This process produces enriched modality representations $\mathbf{F}'_{l,t}$, $\mathbf{G}'$, and $\mathbf{F}'_v$ that reside in a shared embedding space. 
Empirically, the representations exhibit stronger cross-modal alignment compared to the original modality-specific features, indicating effective interaction between the modalities. 

To verify that this block actually learns a shared representation, we measured linear Centred Kernel Alignment (CKA)~\cite{pmlr-v97-kornblith19a} across modality pairs before and after the transformer layer; we present the results in 
Section~\ref{sec:results}. 


\subsubsection{Cross Modal Matching Module}


This module receives three streams of enriched representations from Embodied Fusion Module: updated text tokens $\mathbf{F}'_{l,t}$, gesture tokens $\mathbf{G}'$, and vision tokens $F'_v$, as well as the pointing direction $\hat{\mathbf{d}}_{\text{pred}}$ produced by the Gestural Module. 

\paragraph{Referential-Vision Cross-Attention} 
Text tokens $\mathbf{F}'_{l,t}$ and gesture tokens $\mathbf{G}'$ are concatenated to form a unified referential token sequence $\mathbf{F}_{\text{ref}} \in \mathbb{R}^{(L_t + N_j) \times 128}$, while the updated vision tokens are treated as object-level features. These are passed through a transformer-based cross-modal architecture as described in M3DRef-CLIP~\cite{m3drefclip}, consisting of alternating self-attention blocks augmented with spatial distance weights~\cite{zhao2021_3DVG_Transformer} and cross-attention blocks ($\text{depth}\!=\!2$, $h\!=\!4$), producing a per-proposal content score $S_{\text{conf}} \in \mathbb{R}^{M}$. 

\paragraph{Pointing Bias.}
For the late pointing bias $S_{\text{b}}$, the predicted ray $\hat{\mathbf{d}}_{\text{pred}}$ from the Pointing Regressor of the Gestural Module is used. Given the better-aligned arm (selected by comparing the maximum angular alignment of both arms to the proposals), the cosine similarity between $\hat{\mathbf{d}}_{\text{pred}}$ and the vector from the anchor shoulder ($\mathbf{a}$) to each proposal centre ($\mathbf{c}_i$) is computed in Equation~\ref{eqn:Pointing_Bias}. 

\paragraph{Adaptive Score Fusion.} \label{sec:adaptive_score_fusion}
The content score ($S_{\text{conf}}$) and pointing bias ($S_{\text{b}}$) are combined via:
\begin{dmath}
    S_{\text{final}} = w_0 \cdot S_{\text{conf}} + w_1 \cdot S_b, \quad [w_0, w_1] = \text{Softmax}\bigl(\text{MLP}(\mathbf{F}_{\text{agg}})\bigr),
\end{dmath}
where $\mathbf{F}_{\text{agg}}$ is the global aggregate feature from the Gestural Module's feature extraction branch, reduced to $32$ dimensions and passed through a bias weigher that produces a softmax-normalised two-element weight vector. This mechanism was adopted as is from Ges3ViG. However our results show that this late fusion has little effect on \names as it primarily relies on modality alignment.
This then returns us our final Object Scores, indicating the likely candidate of the reference expression.

\subsection{Losses used and Training Method}
We use the same losses as prior work~\cite{Mane2025Ges3Vig, m3drefclip}: semantic, offset norm, offset direction, and objectness losses from PointGroup, a cross-entropy reference loss, a symmetric contrastive loss, and an L2 joint localisation loss.  
In addition, we introduce a pointing bootstrap loss $\mathcal{L}_{\text{bootstrap}}$, a cosine distance between the regressed pointing direction and the geometric shoulder-to-index ray derived from the model's own predicted joints, applied on first 10 warm-up epochs with linear weight decay, after which it drops to zero; the regressor is trained through the grounding objective. 
The full model is trained end-to-end in a single stage for 40 epochs, with all losses jointly optimised from start.

\subsection{System Implementation}
We now describe our real-world robotic system, which uses \names for 3D-ERU as shown in Figure~\ref{fig:nav_implementation_architecture}. Our system comprises a TurtleBot3 Burger connected to an NVIDIA Jetson Orin Nano, which runs the grounding and navigation pipeline. An Orbbec Femto Bolt depth camera captures a 3D point cloud of the environment, including the human performing the pointing gesture.

Given the point cloud, pointing gesture, and language command, \names predicts a 3D bounding box for the referred object. 
We then select a reachable navigation goal near the predicted target and pass it to Nav2. 
Using a 2D occupancy map derived from the scene point cloud, together with odometry and LiDAR feedback, Nav2 guides the robot toward the selected goal, as shown in supplementary video. 

\section{Experimental Results}\label{sec:results}

\subsection{Dataset \& Evaluation Metrics}
We evaluate on ImputeRefer~\cite{Mane2025Ges3Vig}, which extends ScanNet~\cite{dai2017scannet} and ScanRefer~\cite{chen2020scanrefer} indoor scenes with imputed human avatars pointing at a target object, comprising 35,000 instances across 707 scenes. 
Every instance refers to a single object; the dataset contains no region-level or multi-object references. 
ImputeRefer's referring expressions are shorter and more naturalistic than text-only benchmarks~($\sim10$ words vs. $\sim19$ in ScanRefer), reflecting how language is under-specified when paired with a gesture. This is similar to practical instructions like ``grab that box'' (Figure~\ref{fig:introFig}). Target distance also varies across scenes. 

Following prior work~\cite{Mane2025Ges3Vig}, we use the same definition of \emph{unique} (one object of the target class in the scene) and \emph{multiple} (multiple same-class distractors), and report IoU@0.25 and IoU@0.5 as the primary metrics, along with latency and energy consumed on the Jetson Orin Nano. 


\begin{table}[t]
\centering
\caption{%
  Comparison of \names without foveation against prior 3D-ERU models.
}
\label{tab:model_comparison}
\resizebox{\columnwidth}{!}{%
\setlength{\tabcolsep}{3pt}
\renewcommand{\arraystretch}{0.85}
\small
\begin{tabular}{lcccccc}
\toprule
\multirow{2}{*}{\textbf{Model}}
  & \multicolumn{3}{c}{\textbf{IoU@0.25}}
  & \multicolumn{3}{c}{\textbf{IoU@0.5}} \\
\cmidrule(lr){2-4}\cmidrule(lr){5-7}
  & Unique & Multiple & Overall
  & Unique & Multiple & Overall \\
\midrule
\multicolumn{7}{c}{\textit{without Gestures}} \\
\midrule
3DVG-Transformer \cite{zhao2021_3DVG_Transformer} & 71.56 & 31.35 & 39.17 & 50.66 & 21.54 & 27.20 \\
HAM \cite{chen2023learningpointlanguagehierarchicalalignment} & 67.10 & 25.42 & 33.51 & 48.13 & 16.04 & 22.27 \\
3DJCG \cite{cai20223djcg} & 75.93 & 40.34 & 47.24 & 59.19 & 30.61 & 36.16 \\
RefMask3D \cite{he2024refmask3d} & 72.45 & 25.44 & 34.57 & 64.95 & 22.24 & 30.54 \\
M3DRef-CLIP \cite{m3drefclip} & 77.32 & 62.62 & 65.53 & 60.15 & 47.27 & 49.78 \\
\midrule
\multicolumn{7}{c}{\textit{with Gestures (ImputeRefer)}} \\
\midrule
ScanERU \cite{Lu2024ScanERU}   & 71.6  & 31.91 & 39.54 & 52.79 & 23.06 & 28.84 \\
Ges3ViG \cite{Mane2025Ges3Vig} & 84.6  & 67.6  & 70.9  & 71.0  & 55.8  & 58.7 \\
\textbf{GVT3D w/o Fov.} (Ours)
  & \textbf{86.7} & \textbf{77.4} & \textbf{79.2}
  & \textbf{73.3} & \textbf{65.2} & \textbf{66.7} \\
\midrule
\end{tabular}}
\vspace{-0.25in}
\end{table}

\subsection{Accuracy Comparison with Prior Work on 3D-ERU}

We compare the performance of \names against prior approaches for 3D-ERU, both with and without gestural reasoning, as shown in Table~\ref{tab:model_comparison}. 
Methods listed without Gestures are 3D visual grounding baselines evaluated on the underlying ScanRefer scenes without gesture input; methods listed with Gestures are evaluated on the full ImputeRefer benchmark, including the imputed avatars. 
The results indicate that \names consistently achieves higher accuracy compared to existing state of the art models. 

Among approaches that do not perform gestural reasoning, M3DRef-CLIP achieves the closest performance to \names, but \names outperforms it by 17.7\% in IoU@0.5 on the overall \dataset dataset. 
When compared to Ges3ViG, which supports gestural reasoning but relies primarily on ray tracing for gesture interpretation and lacks cross modal feature alignment, \names achieves an improvement of 8\%.

The improvement is particularly pronounced in the \textit{Multiple} setting, where several instances of the same object class are present in the scene. In this scenario, performance improves from 55.8\% to 65.2\% at IoU@0.5 (+9.4 points). This gain can be attributed to our embodied fusion module, which enables gesture tokens to interact directly with language and geometric features. Such cross modal interaction allows the model to more effectively disambiguate among multiple candidate objects and accurately identify the intended referent. This performance gain highlights the importance of both explicit cross modal feature alignment and moving beyond ray tracing approaches for gestural reasoning.

\subsection{Accuracy, Latency, and Energy Trade-off of \names against Prior 3D-ERU Methods}\label{subsec:accLatEne}
\begin{figure}[t]
    \centering
    \begin{subfigure}[b]{0.49\columnwidth}
        \centering
        \includegraphics[width=\linewidth, height=1.2in]{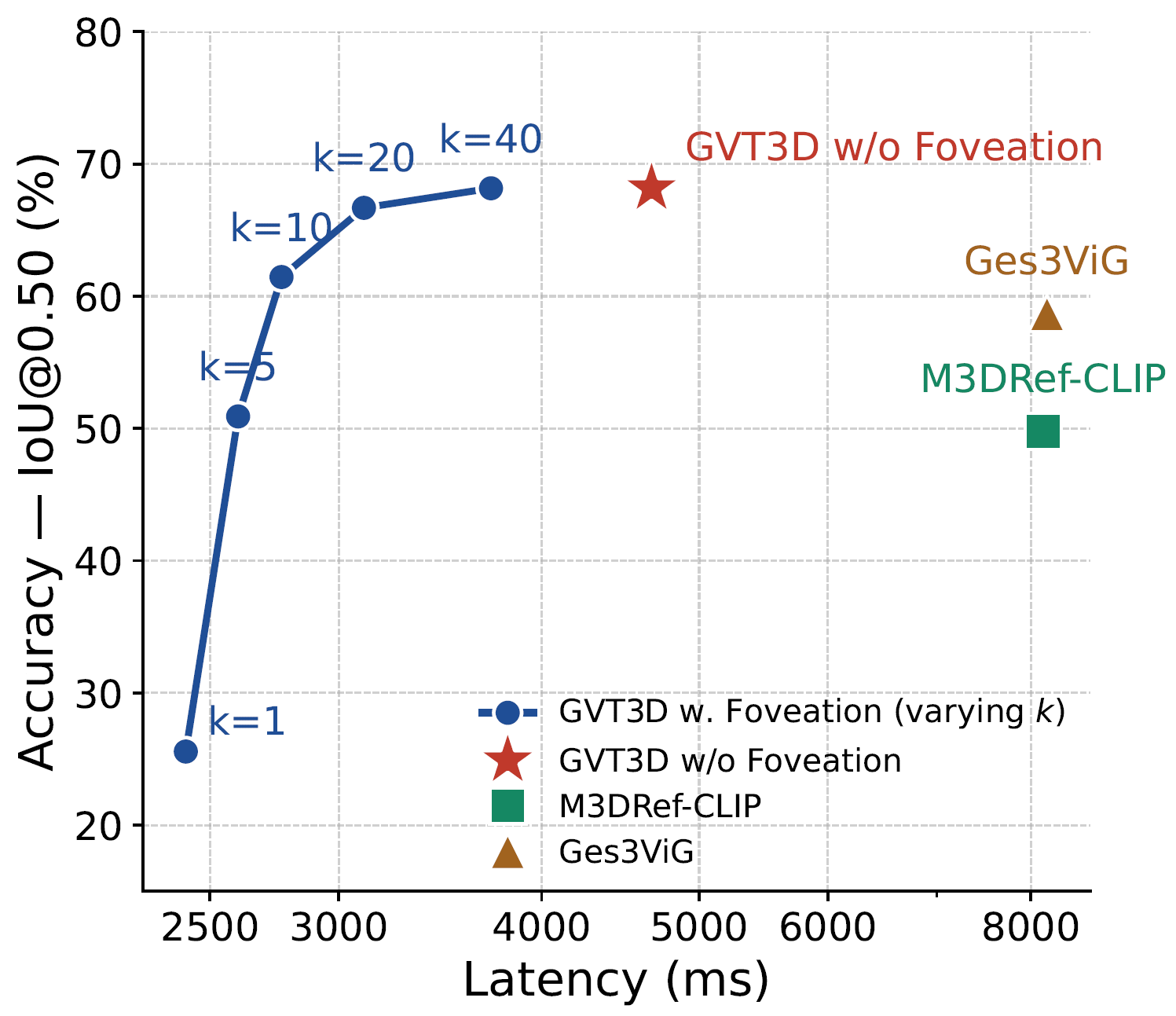}
        \caption{Accuracy vs.\ latency}
        \label{fig:foveation_tradeoff_latency}
    \end{subfigure}
    \hfill
    \begin{subfigure}[b]{0.49\columnwidth}
        \centering
        \includegraphics[width=\linewidth,height=1.2in]{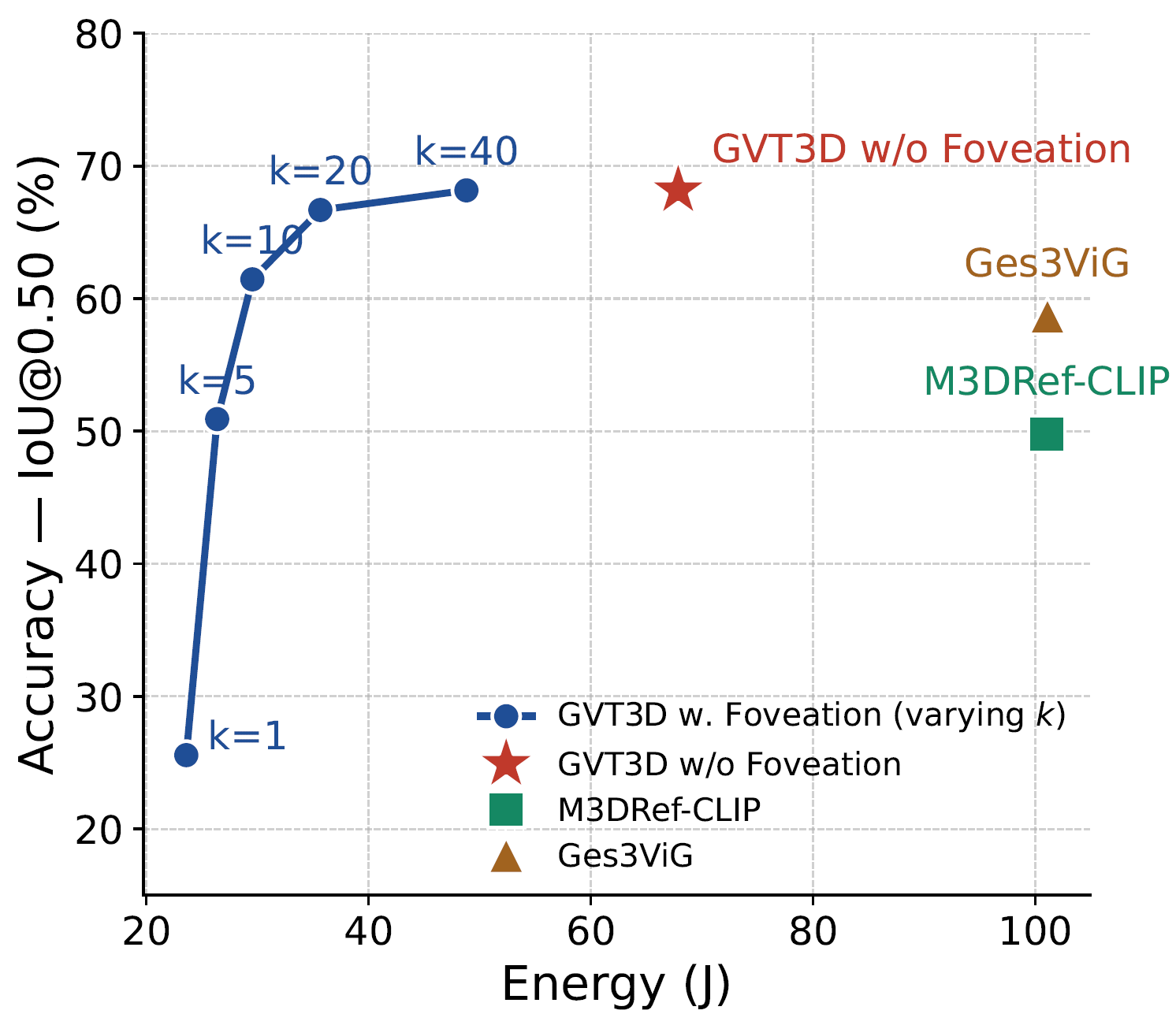}
        \caption{Accuracy vs.\ energy}
        \label{fig:foveation_tradeoff_energy}
    \end{subfigure}
    \caption{Accuracy--latency--energy trade-off of \names{} under varying foveation levels ($k$), measured on the Jetson Orin Nano. \names{} even without applying any foveation remains both faster and more energy efficient than prior methods, while \names with foveation achieving further gains in latency and energy efficiency.}
    \label{fig:foveation_tradeoff}
    \vspace{-0.25in}
\end{figure}

All latency and energy measurements are taken on the Jetson Orin Nano, using 10 warmup scenes to reach a steady operating state, followed by averaging over the same 50 held-out scenes for every model and configuration. 
Figure~\ref{fig:foveation_tradeoff} illustrates the accuracy, latency, and energy trade-off of \names at different $k$ values, where $k$ denotes the maximum number of object proposals retained for foveation, compared with prior 3D-ERU approaches. 
In the accuracy--latency comparison (Figure~\ref{fig:foveation_tradeoff_latency}), \names is 1.75$\times$ faster than Ges3ViG even without foveation, demonstrating its runtime efficiency. Foveation reduces accuracy as $k$ decreases. This occurs mostly in dense scenes, where many objects lie close to the pointing direction. In such scenes, the object nearest to the pointing ray is frequently not the target, because the language instruction identifies a different object in the same region. When $k$ is small, foveation can exclude the correct object from the retained proposals, and the model cannot recover it. However at $k=10$, \names still outperforms Ges3ViG by 2.75\% in IoU@0.5 while running 2.95$\times$ faster.
Similarly, in the accuracy--energy comparison (Figure~\ref{fig:foveation_tradeoff_energy}), \names without foveation achieves higher accuracy while consuming only $67.11\%$ of the energy required by Ges3ViG. With foveation, \names maintains higher accuracy than Ges3ViG down to $k=10$, where it consumes only \textbf{$29.24\%$} of Ges3ViG's energy. These results demonstrate that \names achieves higher accuracy with lower latency and energy consumption than Ges3ViG, supporting its suitability for on-device ERU tasks in robotic deployments.


\begin{table}[t]
\centering
\caption{%
 Ablation studies for \names with \dataset dataset
}
\label{tab:model_ablations}
\resizebox{0.85\columnwidth}{!}{%
\setlength{\tabcolsep}{3pt}
\renewcommand{\arraystretch}{0.85}
\small
\begin{tabular}{lcccccc}
\toprule
\multirow{2}{*}{\textbf{Configuration}}
  & \multicolumn{3}{c}{\textbf{IoU@0.25}}
  & \multicolumn{3}{c}{\textbf{IoU@0.5}} \\
\cmidrule(lr){2-4}\cmidrule(lr){5-7}
  & Unique & Multiple & Overall
  & Unique & Multiple & Overall \\
\midrule
w/o Fusion  & 80.1 & 59.0 & 63.2 & 68.0 & 47.8 & 51.6 \\
w/o Ges.    & 65.9 & 27.3 & 34.8 & 57.4 & 22.2 & 29.0 \\
w. fixed lang.   & 71.7 & 68.9 & 69.3 & 63.1 & 58.4 & 59.3 \\
Ges3ViG \cite{Mane2025Ges3Vig} (w. fixed lang) & 51.1 & 44.9 & 46.1 & 43.9 & 36.8 & 38.2 \\
\midrule
\textbf{GVT3D w/o Fov.} (Ours)
  & \textbf{86.7} & \textbf{77.4} & \textbf{79.2}
  & \textbf{73.3} & \textbf{65.2} & \textbf{66.7} \\
\bottomrule
\end{tabular}}
\vspace{-0.05in}
\end{table}

\subsection{Ablation Studies: } 
Table~\ref{tab:model_ablations} presents several ablation studies of \names on the \dataset dataset. 
In the w/o Fusion setting, we remove the embodied fusion module. Under this configuration, we observe a significant degradation in performance, with IoU@0.5 on the overall dataset dropping by 15.9\%. This highlights the importance of cross modal feature alignment. 

In the w/o Ges. setting, the human is removed from the scene environment, forcing \names to rely entirely on language based reasoning. In this, we again observe a substantial reduction in performance, with IoU@0.5 decreasing by 38.5\% on the overall dataset. This degradation is even more pronounced in the \textit{Multiple} setting, where the performance drops by 43.5\%. This result indicates that gestural reasoning plays a critical role in disambiguating between objects of the same class when multiple candidates are present. 
Finally, in the w. fixed lang. setting, where the language description for the entire dataset is replaced with a minimal expression `this one', we observe that \names achieves 20\%  higher accuracy when compared to Ges3ViG.
Collectively, these ablation results demonstrate the importance of accurate cross modal fusion, as well as the complementary roles of language and gesture reasoning in 3D-ERU.



\subsection{Evaluation of Embodied Fusion Module}

\begin{figure}[t]
    \centering
    \begin{minipage}{0.47\linewidth}
        \centering
        \includegraphics[width=\linewidth]{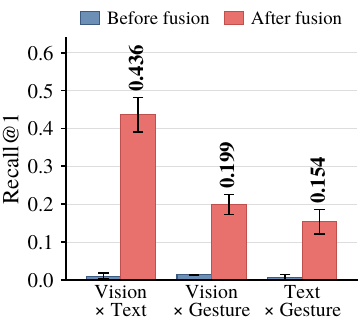}
        \caption{Cross-modal Recall@1 before and after fusion across modality pairs ($200$ val samples).}
        \label{fig:recall_bars_3wayca_full}
    \end{minipage}
    \hfill
    \begin{minipage}{0.47\linewidth}
        \centering
        \includegraphics[width=\linewidth]{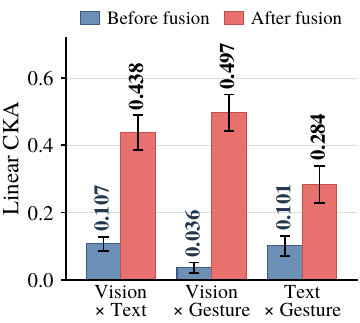}
        \caption{Linear CKA between modality pairs before and after the embodied fusion block.}
        \label{fig:cka_bars_full}
    \end{minipage}
    \vspace{-0.25in}
\end{figure}

\begin{figure}[t]
    \centering
    \includegraphics[width=0.75\columnwidth]{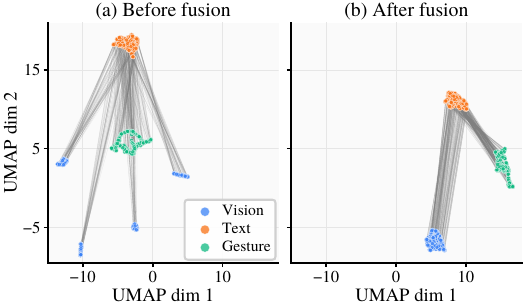}
    \caption{UMAP of vision, text, and gesture embeddings before (a) and after (b) fusion. Lines connect matched triplets. After fusion, clusters converge and connecting lines become parallel, reflecting consistent cross-modal alignment.}
    \label{fig:tsne_sample_allign}\vspace{-0.1in}
\end{figure}

To verify cross-modal alignment, we measure Linear CKA \cite{pmlr-v97-kornblith19a} and Recall@1 before and after the fusion block. As shown in Figures~\ref{fig:recall_bars_3wayca_full} and~\ref{fig:cka_bars_full}, all modality pairs exhibit large gains: CKA increases by $+0.33$, $+0.46$, and $+0.18$ for vision-text, vision-gesture, and text-gesture respectively, while Recall@1 improves from near-zero to $43.6\%$, $19.9\%$, and $15.4\%$. 
This is further confirmed by Figure~\ref{fig:tsne_sample_allign} (UMAP projections):~initially separated modality clusters draw together after the fusion, with matched triplets forming tight local neighborhoods.

We additionally measured the causal contribution of the language and gesture branches to the final score under various localized sensor degradations, as shown in Figure~\ref{fig:modality_noise}. For each stream we zero it before the fusion block, replay the model's own fusion, matching and score-blending head, and record the drop $\Delta z_m$ in the z-score of the originally selected proposal within the score distribution over valid proposals; rectifying and normalizing the three drops gives a decision share $s_m = [\Delta z_m]_+ / \sum_{m'} [\Delta z_{m'}]_+$ that sums to one over language, gesture and vision. Under noise localized to the target, gesture reliance increases as visual evidence degrades. Conversely, when noise is applied around the human, the gesture contribution more than halves while the language share increases from approximately $31\%$ to $44\%$. This shows the \names can adapt the importance of gesture and language according to context. 

\subsection{Human-to-Target Distance-based Sensitivity Study}

\begin{figure}[t]
    \centering
    \begin{minipage}[t]{0.47\columnwidth}
        \centering
        \includegraphics[width=\linewidth]{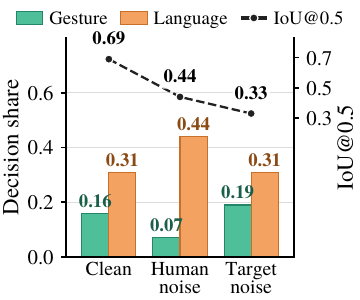}
        \caption{Causal decision share per modality under $\sigma = 2$\,cm localized point-cloud noise, with IoU@$0.5$.}
        \label{fig:modality_noise}
    \end{minipage}
    \hfill
    \begin{minipage}[t]{0.47\columnwidth}
        \centering
        \includegraphics[width=\linewidth]{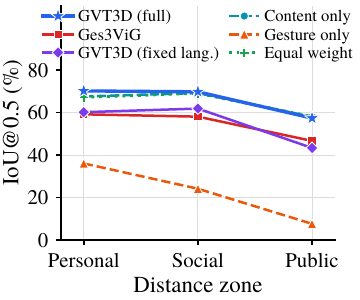}
        \caption{Sensitivity scan for different distances: Personal, Social, Public.}
        \label{fig:distance_sensitivity}
    \end{minipage}
    \vspace{-0.2in}
\end{figure}

To understand pointing gesture effectiveness variance with distance, we analyze model performance across proxemic zones~\cite{hall1966hidden}: personal (0.46--1.22m), social (1.22--3.70m), and public ($\geq$3.70m), shown in Figure~\ref{fig:distance_sensitivity}. Performance stays stable at personal (70.1\%) and social (69.7\%) zones but drops at public zone (57.3\%), where gesture cues become less discriminative.
We also evaluate three fixed-weight variants~\ref{sec:adaptive_score_fusion}: \textit{content only} ($w_0{=}1, w_1{=}0$), \textit{gesture only} ($w_0{=}0, w_1{=}1$), and \textit{equal weight} ($w_0{=}w_1{=}0.5$). Content only ($S_{conf}$) performs near-identically to the full model, confirming that the model genuinely learns from all three modalities rather than relying on ray-casting. Gesture only ($S_b$) degrades significantly across all zones, demonstrating that raw pointing rays alone are insufficient for grounding.

These retrieval results confirm that the fusion block enables matched samples from different modalities to locate one another in embedding space, demonstrating that the module aligns the three modalities rather than merely mixing features indiscriminately.

\subsection{Success Rate Evaluation with \names}


We evaluated \names with our robotic prototype in real-world scenes that were not part of the training data with our system as shown in Figure~\ref{fig:nav_implementation_architecture}. We placed the target between 2.75 m and 9.5 m from the robot across the trials (mean 5.89 m, SD 2.28 m). Before these trials, we fine-tuned the model on 80 samples that we collected from real persons, and we replaced the synthetic persons in the \dataset dataset with these samples.

Two sources of error can occur in this pipeline: (i)~\names grounds the wrong object, resulting in an incorrect navigation goal, and (ii)~the TurtleBot fails to reach the correct goal due to navigation drift or failure.
To assess (i), six participants each viewed 30 scenes showing the real-world environment and the predicted target bounding box, and judged whether the grounding was correct. Overall, \namef object localization was judged correct in 70\% of instances, as compared to 56\% of cases in Ges3ViG. 
To assess (ii), we evaluated navigation performance on the subset of trials where \names grounded the correct object. Across 16 completed trials, the TurtleBot reached within a mean distance of 0.79~m (SD 0.36~m) from the target. Navigation was aborted in 2 trials due to odometry drift.


\section{Conclusion}\label{sec:conclusion}

We propose \name, a novel approach for 3D-ERU that localises objects in 3D scenes given language and pointing gestures. 
Unlike prior methods that rely on heuristic ray tracing for gestural reasoning, \names models gestures as a spatial prior jointly conditioned with language features. 
We introduce an embodied fusion mechanism that tightly aligns vision, language, and gesture representations within a shared embedding space. 
Experimental results show that (a) \names achieves $\sim$8\% higher accuracy without foveation and (b) foveation enables 2.95$\times$ faster execution while maintaining comparable accuracy, relative to prior 3D-ERU models. 

\section{Acknowledgments}

This research was supported by the Singapore National Research Foundation (NRF), Prime Minister’s Office, Singapore, under its Campus for Research Excellence and Technological Enterprise (CREATE) programme. The Mens, Manus, and Machina (M3S) is an interdisciplinary research group (IRG) of the Singapore-MIT Alliance for Research and Technology (SMART) centre.\bibliographystyle{IEEEtran}
\bibliography{ref}

\end{document}